# Amplifying Aspect-Sentence Awareness: A Novel Approach for Aspect-Based Sentiment Analysis

Adamu Lawan[1], Juhua Pu[1], Haruna Yunusa[2], Jawad Muhammad[3], Aliyu Umar[4]

[1] School of Computer Science and Engineering, Beihang University, Beijing, China
[2] School of Automation Science and Electrical Engineering, Beihang University, Beijing, China
[3] School of Artificial Intelligence, University of Chinese Academy of Sciences, Beijing, China
[4] Jigawa State Institute of Information Technology, Kazaure

**Abstract.** Aspect-Based Sentiment Analysis (ABSA) has gained prominence as a natural language processing (NLP) technique for extracting and analyzing sentiments related to specific aspects or features within text, offering detailed insights from sources like customer reviews and social media feedback. However, existing attention-based models often struggle to effectively connect aspect with context due to language complexity and multiple sentiment polarities in a single sentence. To address these challenges, we propose Amplifying Aspect-Sentence Awareness (A3SN), a novel technique designed to enhance ABSA by amplifying aspect-sentence awareness. Our approach follows the transformer's standard process and incorporates multi-head attention (MHA) mechanisms to enrich the model with the input text's semantic information. We introduce an enhanced MHA mechanism that amplifies aspect-sentence awareness attention, doubling the attention between the sentence and the aspect to capture subtle relationships and dependencies accurately. Furthermore, a gated fusion mechanism was used to integrate feature representations from the multi-head and amplify aspect-sentence awareness attention modules. Experimental results on three benchmark datasets demonstrate A3SN's effectiveness, outperforming state-of-the-art (SOTA) baseline models while maintaining simplicity and low computational complexity.

# 1    Introduction

In Natural Language Processing (NLP), text classification is fundamental in understanding and extracting insights from textual data. Sentiment analysis, a crucial component of text classification, involves determining the emotional tone or sentiment expressed in a text. With the rapid increase of online platforms and the abundance of user-generated content, sentiment analysis has gained increasing importance in various applications such as customer feedback analysis and product recommendation systems. However, traditional sentiment analysis often provides a broad overview of sentiment without considering the specific aspects or features mentioned in the text. This limitation led to the development of ABSA, which aims to identify overall sentiment and extract and analyze sentiments related to specific aspects or features mentioned in the text. ABSA enhances the detail of sentiment analysis, providing valuable insights for informed decision-making and improving user experiences. For instance, consider the sentence: "The camera quality of this smartphone is excellent, but the battery life is disappointing." Here, ABSA identifies the overall sentiment (positive towards the camera, negative towards battery) and specific aspects being praised or criticized. This detailed insight is invaluable for businesses to address concerns, improve products, and tailor marketing strategies effectively.

In the field of ABSA, attention-based (semantic) approaches [1], [2], [3], [4], [5], [6], [7], [8], [9], [10], [11] have emerged as powerful tools for unraveling the intricate aspects sentiments expressed within texts. [7] pioneered deep memory networks, shedding light on the significance of individual context words in aspect-level sentiment classification. Building upon this foundation, [2] introduced attention-based extended short-term memory networks, emphasizing the relationship between sentiment polarity and specific sentence aspects. [6] further refined this approach with interactive attention networks, recognizing the importance of modeling targets and contexts separately to enhance sentiment classification precision. Meanwhile, [3] explored the landscape of ABSA with a novel neural network-based framework leveraging multiple-attention mechanisms. Their integration of recurrent neural networks and weighted-memory mechanisms bolstered the model's ability to handle complex ABSA tasks. [9] the multi-grained attention network was introduced to address information loss with multi-word aspects by combining fine-grained and coarse-grained attention mechanisms. As the journey continued, attention mechanisms remained at the forefront of innovation. [4] proposed enhancements to attention effectiveness, integrating syntactic information into the attention mechanism to refine target representation. [8] introduced MHA networks, emphasizing robust interactions between context and aspect terms to improve ABSA performance. [12] introduced attentional encoder networks as a more efficient alternative to recurrent neural networks, offering improved sentiment classification toward specific targets. Using a multi-attention network and global and local attention modules to capture differentially grained interacting information between aspect and context, [1] overcame the shortcomings of previous approaches.

In parallel, advancements in graph-based (syntactic) approaches [13], [14], [15], [16], [17], [18], [19] reshaped the landscape of ABSA. [13] proposed merging convolution over a dependency tree (CDT) with bi-directional long short-term memory (Bi-LSTM) to analyze sentence structures effectively. Additionally, it introduces graph convolutional network (GCN) enhancements, directly engaging with sentence dependency trees. [14] tackled the challenge of connecting aspects with opinion words using relational graph attention networks, enabling more accurate sentiment prediction. Transitioning to recent developments, [20] introduced type-aware graph convolutional networks (T-GCN) for ABSA, explicitly considering dependency types to enhance model performance. Meanwhile, [16] addressed the challenges of multiple-aspect sentiment representation with heterogeneous graph neural networks, providing a unified framework for encoding sentence syntax, word relations, and opinion dictionary information. Looking ahead, [17] introduced KHGCN, a knowledge-guided heterogeneous graph convolutional network that maximizes BERT's potential and dynamically merges subword vectors to enhance sentiment analysis accuracy. Concurrently, [19] proposed ASGCN to capture comprehensive semantic information for aspect-level sentiment analysis by incorporating grammatical dependencies and semantic connections. In response to the limitations of existing neural networks, [21] introduced SGAN, a syntactic graph attention network that includes knowledge of dependency types to improve the propagation of word representations. Similarly, [18] addressed the challenges of inter-word relationships and syntactic dependencies with ASHGAT, leveraging a word-level relational hypergraph to optimize hypergraph attention for enhanced sentiment classification accuracy.

Because of their frequent inability to comprehend the intricate interaction between aspect terms and context, existing attention-based models have trouble identifying overlapping features when expressing various sentiment polarities. Recent research graph-based models have highlighted the advantages of leveraging syntactic information, such as dependency trees, to capture long-range syntactic relationships effectively and the connections between aspect and context. However, these approaches face challenges, including vulnerability to parsing errors, where minor inaccuracies can significantly impact the model's performance. Furthermore, integrating syntactic structure with semantic correlations can increase model complexity and introduce additional computational overhead. In summary, it is essential to address challenges such as under-exploring the relationship between aspect terms and context and computational complexity to enhance the robustness and efficiency of ABSA models.

To effectively address the above problems, in this work, we present A3SN, a novel approach aimed at improving the model's comprehension of the subtle relationship between aspect and sentence in ABSA. This method comprises an MHA mechanism to enrich the model with semantic information from the input text. It also includes an amplify aspect-sentence awareness attention, which enhances the aspect-sentence relationship. By doubling the attention directed toward the aspect and the sentence within the input sequence, our approach effectively emphasizes its significance, enabling the model to capture subtle relationships and dependencies more accurately. Lastly, a gated fusion is introduced to adaptively integrate feature representations from the MHA and aspect-sentence awareness attention module, ensuring the selection and aggregation of valuable information essential for ABSA. Overall, A3SN enhances the model's capability to understand and analyze sentiment within the context of specific aspects.

The main contributions of this paper are as follows:
- To the best of our knowledge, we are the first to investigate the impact of enhancing aspect-sentence awareness attention in ABSA.

- Amplify aspect-sentence awareness attention enhances the MHA mechanism by doubling the attention on aspect-sentence relationships. This enhancement helps the model capture subtle relationships and dependencies more accurately.
- A3SN enhances sentence-aspect relationships in ABSA by using MHA, amplify aspect-sentence awareness attention and gated fusion on top of BERT. This approach captures complex dependencies and improves ABSA performance.
- The experimental results on three benchmark datasets (Restaurant14, Laptop14, and Twitter) showcase the effectiveness of the A3SN model, surpassing SOTA baseline models that incorporate semantic, syntactic, and common knowledge.

## 2    Related Work

In ABSA, methods related to our study fall into two categories: attention-based models focusing on semantics and a combination of attention-based semantic models with graph-based syntactic models, which we can also refer to as hybrid models.

### 2.1    Attention-based Models

Recent studies in aspect-based sentiment analysis primarily address the task by employing attention-based neural networks. These models are designed to capture the semantic relationship between the context and the aspect term, enabling a deeper understanding of sentiment expressed towards specific aspects within the text [1], [2], [3], [4], [5], [6], [7], [8], [9], [10], [11].

[7] proposed a deep memory network for aspect-level sentiment classification, emphasizing the importance of individual context words in sentiment classification. In this approach, neural attention models were incorporated with external memory to adeptly capture subtle variations in sentiment expression. [2] focused on the relationship between sentiment polarity and specific aspects within sentences. They introduced an attention-based extended short-term memory network (LSTM) tailored for aspect-level sentiment classification, employing attention mechanisms to highlight distinct

sentence parts based on different aspects. [6] addressed the separate modeling of targets and contexts in sentiment classification, proposing interactive attention networks (IAN). IAN facilitated interactive learning by leveraging interactive attention mechanisms and generated distinct representations for targets and contexts, enhancing sentiment classification precision. [3] presented a neural network-based framework for identifying sentiment towards opinion targets in comments or reviews. Their approach leveraged a multiple-attention mechanism to capture sentiment features across long distances, integrating multiple attentions with a recurrent neural network for improved expressiveness. [9] introduced a multi-grained attention network (MGAN) for aspect-level sentiment classification. Unlike existing approaches, MGAN utilized fine-grained attention mechanisms to capture word-level interactions between aspect and context, enhancing classification accuracy. [4] proposed enhancements to attention effectiveness were made by refining target representation and integrating syntactic information into the attention mechanism, addressing challenges in capturing the relationship between aspect terms and context. [22] introduced conditional BERT contextual augmentation, a data augmentation technique specifically designed for labeled sentences in NLP. This approach enhances data diversity by replacing words with varied alternatives predicted by a language model, leveraging recent advancements in contextual augmentation. Furthermore, the authors retrofit BERT into a conditional masked language model, known as conditional BERT, which substantially improves the quality and diversity of augmented data. Overall, their method offers a promising solution for enhancing model generalization and reducing overfitting in deep learning tasks. [8] addressed limitations in previous attention-based ABSA methods by proposing a novel approach centered on MHA networks. Leveraging pre-trained embeddings from BERT, combined with MHA and convolutional operations, their approach improved context-aspect interactions and enhanced ABSA performance. [1] presented a multi-attention network (MAN) addressing various challenges by employing intra-level attention with a transformer encoder and inter-level attention with global and local modules. [5] addressed the challenge of ABSA by proposing semantic distance Attention with the BERT model (SDA-BERT), using BERT to extract aspect semantic features for improved sentiment analysis accuracy. [23] presented a new model for analyzing senti-ment at the aspect level, utilizing bidirectional encoder representations from transformers (BERT) to create word embeddings. They produced sentence representations using various attention strategies, including intra- and inter-level attention. MHA and point-wise feed-forward structures are examples of intra-level attention. In contrast, inter-level attention uses global attention to capture interactions between aspect and context words. Furthermore, to improve sentiment analysis, a feature-focused attention technique is presented. [10] proposed a hybrid network model for aspect-level sentiment classification, incorporating positional and interactive MHA mechanisms to capture multi-level emotional characteristics and discern subtle sentiment patterns in complex sentences. [11] introduced the interactive MHA capsule network model (IMHSACap) to enhance sentiment analysis performance. By prioritizing local context information and capturing long-range dependencies between global and local contexts, IMHSACap improved aspect-level sentiment analysis accuracy.

## 2.2 Hybrid-based Models

Various studies in recent years have explored the combination of attention mechanisms, common knowledge, graph convolutional networks (GCNs), or graph attention networks (GATs) to enhance ABSA. [14] developed a unified aspect-oriented dependency tree structure and proposed the relational graph attention network (R-GAT) to improve sentiment prediction accuracy by focusing on relevant sentence aspects. [15] addressed limitations in existing graph-based ABSA approaches by proposing BERT4GCN, integrating sequential features from BERT and syntactic knowledge from dependency graphs for improved classification performance. [20] introduced type-aware graph convolutional networks (T-GCN) for ABSA, explicitly considering dependency types to discern different edges in the graph and effectively learn from various layers. [24] proposed using a novel gated graph convolutional network on the dependency tree to encode syntactical information and a syntax-aware context dynamic weighted layer to enhance focus on local syntax-aware context, with MHA capturing both semantic information and the interaction between semantics and syntax. [16] discussed problems in aspect-level sentiment classification and proposed heterogeneous graph neural networks (Hete_GNNs) to encode sentence syntax trees, word relations, and opinion dictionary information, improving sentiment prediction. [25] AG-VSR integrates two key representations for classification: attention-assisted graph-based representation (A2GR) and variational sentence representation (VSR). A2GR, generated by a GCN module processing a modified dependency tree, and VSR, obtained from a learned distribution via an encoder-decoder structure akin to a variational autoencoder, collectively enhance the model's ability for aspect-based sentiment analysis. Recent developments include KHGCN [17], ASGCN [19], SGAN [21], and ASHGAT [18], each introducing novel approaches to aspect-based sentiment analysis by leveraging advanced graph neural network architectures and attention mechanisms tailored to specific aspects and context relationships.

### 2.3   Discussion

Existing attention-based models in ABSA struggle to fully grasp the intricate relationship between aspect terms and context, making it challenging to recognize overlapping features when expressing multiple sentiment polarities. Recent research suggests leveraging syntactic information, like dependency trees, to capture long-range syntactic relationships and aspect-context connections effect. However, these approaches face challenges, including vulnerability to parsing errors and increased model complexity. Addressing these challenges is crucial for enhancing the robustness and efficiency of ABSA models.

## 3     Overview of our Proposed Model Framework

In the ABSA task, we define it as a sequence-to-class problem. Given a sentence-aspect $(s, a)$, where $s = \{w_1, w_2, ..., w_n\}$ represents the tokens in the sentence, and $a = \{a_1, a_2, .., a_m\}$ is a sub-sequence of s denoting the aspect, our objective is to determine the sentiment polarity of aspect a within sentence s.

We introduce a novel technique called A3SN to improve the model's understanding of the complex relationship between the aspect and the sentence. This innovative approach incorporates a textual semantic module, where MHA weights are computed following the transformer's standard process [26]. This module serves as a channel to enrich the model with sentence and aspect information. In addition, we modify the MHA mechanism by amplifying the attention weights in the aspect-sentence awareness attention module, by doubling the attention specifically between the sentence and the aspect, within the input sequence. This amplification of attention directed towards the aspect effectively highlights its importance within the context of the entire sentence. We also use the gated fusion, an adaptively integrated feature representation system combining the amplified attention and textual semantic representations. As a filter at the end of the two modules, gated fusion gathers and chooses important data necessary for the ABSA task. This comprehensive approach significantly enhances the model's ability to understand and analyze sentiment within specific aspects. Our method also enables the model to accurately capture the subtle relationship and dependencies between the aspect and the sentence.

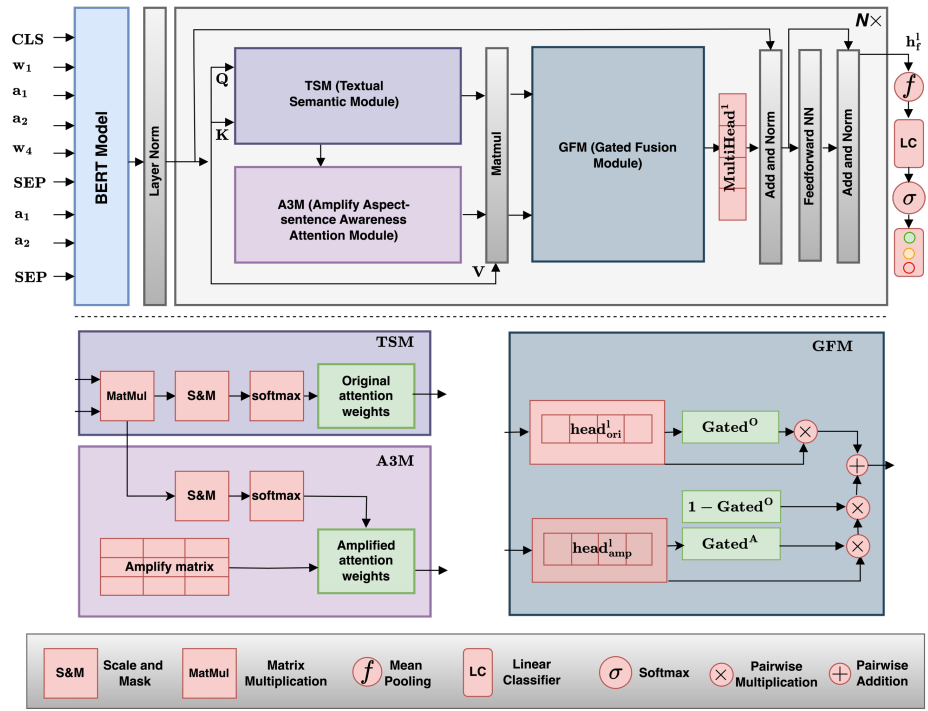

**Fig. 1.** A3SN complete architecture

### 3.1  Embedding Module

In BERT [27] encoding, the sentence-aspect is structured as '[CLS]+sentence +[SEP]+aspect+[SEP]', forming the input sequence [12]. This format allows for extracting an aspect-aware hidden state vector, denoted as *h*. This aspect-aware hidden state vector serves as a rich representation that incorporates information from both the input sentence and the associated aspect, enabling more understanding and analysis in ABSA tasks.

### 3.2  Textual Semantic Module

To extract important textual semantic information related to the given sentence and aspect, we employ the MHA mechanism called the "Textual semantic module" as shown in Fig. 1. In the textual semantic module, computation adheres to the standard process of transformer architecture. The first step in computing attention weights $score_{ori}^l$ is to take the dot product of the keys $K^l$ and queries $Q^l$. Next, another dot product between and the values $V^l$ yields the output representation $head_{ori}^l$ of the attention module. Below is an outline of this method:

$$K^l, Q^l, V^l = LN(h^l)W_k, LN(h^l)W_q, LN(h^l)W_v \tag{1}$$

$$score_{ori}^l = \frac{softmax(Q^l K^{l^T} + Mask)}{\sqrt{d_k}} \tag{2}$$

$$head_{ori}^l = score_{ori}^l V^l \tag{3}$$

The trainable parameters $W_v, W_k$, and $W_q$ are used to calculate the values, keys, and queries of the $l^{th}$ layer, which are represented by $K^l$, $Q^l$, and $V^l$, respectively. *LN* is referred to as layer normalization. A pre-trained masking language modeling matrix is represented by *M*ask, and the dimension of $K^l$ is indicated by $d_k$. $Score_{ori}^l$ denotes the attention weight matrix sized $N \times N$, where each entry determines the weight assigned to a token in the sentence and aspect for computing the representation of another token. The attention mechanism encompasses two types of attention within $score_{ori}^l$ as shown in Fig. 2: intra-attention (*attention* between sentence (s, s), and *attention* between aspect (a, a)) and aspect-sentence awareness attention (*attention* between sentence and aspect (s, a) and *attention* between aspect and sentence (a, s)). This delineates how tokens within the sentence or aspect and across aspect and sentence are weighted in the computation process, facilitating contextual understanding in the model's representation generation.

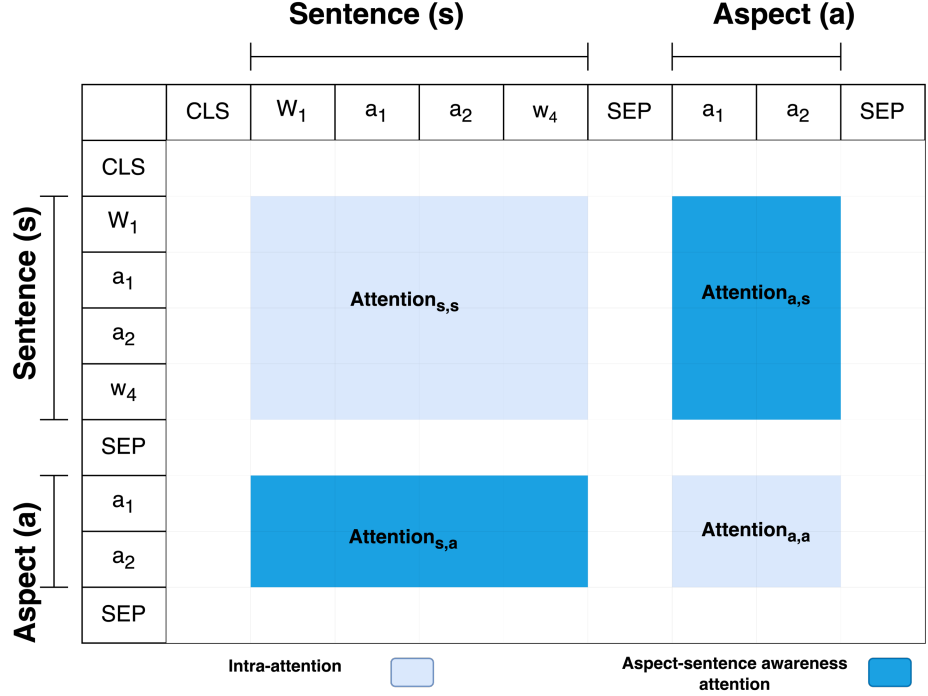

Fig. 2. The vanilla attention weights, where $attention_{s,s}$ and $attention_{a,a}$ in grey represent intra-attention and $attention_{s,a}$ and $attention_{a,s}$ in blue represent aspect-sentence awareness attention

### 3.3   Amplify Aspect-Sentence Awareness Attention Module

The inspiration for our proposed architecture stems from the extensive validation of the MHA mechanism within transformer models. Studies have consistently demonstrated the meaningfulness of the learned attention weights, underlining their pivotal role in capturing intricate linguistic relationships. Building upon this foundation, we introduce a simple yet impactful strategy: doubling the attention allocated to the sentence-aspect interaction. By doing so, we aim to encourage the model to extract and harness richer aspect-sentence information. This augmentation is achieved by utilizing an amplify matrix, denoted as $Amplify_{mat}$, which mirrors the size of the multi-head attention weight matrix, as shown in Fig. 1. Through this approach, we seek to enhance the model's ability to comprehend and leverage the inherent relationship between sentences and aspects, ultimately improving its performance in ABSA tasks. Given the final input sequence [CLS] $w_1$, $a_1$, $a_2$, $w_4$ [SEP] $a_1$, $a_2$ [SEP], $Amplify_{mat}$ is formulated as:

$$Amplify_{mat} = \begin{pmatrix} g_{1,1} & g_{1,2} & \cdots & g_{1,N} \\ g_{2,1} & g_{2,2} & \cdots & g_{2,N} \\ \vdots & \vdots & \ddots & \vdots \\ g_{N,1} & g_{N,2} & \cdots & g_{N,N} \end{pmatrix} \quad (4)$$

where the elements $g_{i,j}$ are defined as:

$$g_{i,j} = \begin{cases} 2, & \text{if } w_i \in s \text{ and } w_j \in a \text{ or } w_i \in a \text{ and } w_j \in s \\ 1, & \text{otherwise} \end{cases} \quad (5)$$

The amplified attention score $Score_{amp}^l$ is then computed as the element-wise multiplication (Hadamard product) of the original attention score $score_{ori}^l$ and the amplified matrix ($Amplify_{mat}$).

$$Score_{amp}^l = score_{ori}^l \circ Amplify_{mat} \quad (6)$$

$$Score_{amp}^l = Amplified(score_{ori}^l) \quad (7)$$

$$head_{amp}^l = Score_{amp}^l V^l \quad (8)$$

Where Amplified(.) represents the amplification procedure.

This method amplifies the model's focus on sentence-aspect interactions by assigning higher weights to relevant token pairs. As a result, the model gains improved capability in extracting and incorporating pertinent sentence-aspect information, leading to enhanced performance in ABSA.

### 3.4 Gated Fusion Module

Gated fusion has demonstrated effectiveness in language modeling tasks [28]. To dynamically assimilate valuable insights from both the textual semantic module representation and the amplify aspect-sentence awareness attention module representation, we used a gated fusion module to reduce interference from unrelated data. Gating is a potent mechanism for assessing feature representations utility and orchestrating information aggregation accordingly. This module uses a simple addition-based fusion mechanism to achieve gating, which controls the flow of information through gate maps as shown in Fig. 1. Specifically, the representations $head_{ori}^l$ and $head_{amp}^l$ are associated with gate maps $Gate^O$ and $Gate^A$, respectively. These gate maps originate from a convolutional neural network (CNN) using a one-dimensional convolutional layer. These gate maps are used to provide technical specifications for the gated fusion process:

$$Gate^O = \sigma(CNN(head_{ori}^l)) \quad (9)$$

$$Gate^A = \sigma(CNN(head_{amp}^l)) \quad (10)$$

$$head^l = Gate^O head_{ori}^l + (1 - Gate^O)Gate^A head_{amp}^l \quad (11)$$

Each attention head's representation is combined to create the multi-head attention mechanism's output representation, $MultiHead$. This concatenation operation is expressed as:

$$MultiHead^l = concat(head_1^l, \ldots, head_m^l)w_h \quad (12)$$

$$\tilde{h}_f^l = LN(LN(h^l) + MultiHead^l) \quad (13)$$

$$h_f^l = LN(FFN(\tilde{h}_f^l) + \tilde{h}_f^l) \quad (14)$$

where the linear layer's parameters are represented by $w_h$, and $m$ stands for the number of attention heads. This process ensures that the model effectively integrates insights from all attention heads to generate the final representation, optimizing its performance for the given task.

We employ mean pooling to condense contextualized embeddings $h_f^l$, which assists downstream classification tasks. Following this, we apply a linear classifier to generate logits. Finally, softmax transformation converts logits into probabilities, facilitating ABSA. Each component is pivotal in analyzing input text for ABSA tasks from the embedding layer to the sentiment classification layer.

$$h_{mp}^l = MeanPooling(h_f^l) \quad (15)$$

$$p(a) = softmax(W_p h_{mp}^l + b_p) \quad (16)$$

$W_p$ and $b_p$ represents the trainable parameters, consisting of learnable weights and biases.

### 3.5   Training

We utilize the standard cross-entropy loss as our primary objective function:

$$L(\theta) = -\sum_{(s,a)\in D} \sum_{c\in C} \log p(a) \qquad (17)$$

computed over all sentence-aspect pairs in the dataset $D$. For each pair $(s, a)$, representing a sentence $(s)$ with aspect $(a)$, we compute the negative log-likelihood of the predicted sentiment polarity $p(a)$. Here, $\theta$ encompasses all trainable parameters, and $(C)$ denotes the collection of sentiment polarities.

## 4      Experiment

### 4.1   Datasets

Our experiments utilize three public sentiment analysis datasets: the Laptop and Restaurant14 review datasets from the SemEval 2014 Task[29] the Twitter dataset employed by [30]. For detailed statistics of these datasets, refer to Table 1.

**Table 1.** Statistics of three benchmark datasets

| Dataset | Division | Positive | Negative | Neutral |
|---|---|---|---|---|
| Restaurants14 | Train | 2164 | 807 | 637 |
|  | Test | 727 | 196 | 196 |
| Laptop14 | Train | 976 | 851 | 455 |
|  | Test | 337 | 128 | 167 |
| Twitter | Train | 1507 | 1528 | 3016 |
|  | Test | 172 | 169 | 336 |

### 4.2   Implementation

Our A3SN model uses the pre-trained BERT model to extract word representations from the last hidden states [27]. We adopt an MHA mechanism with 4 heads for enhanced representation learning. For the model architecture, we experimented with varying numbers of layers. Three layers proved optimal for the Laptop and Twitter datasets, whereas the Restaurant dataset achieved superior performance with just one layer. These representations are fine-tuned during training to adapt to our specific task. We implement the model using the PyTorch framework, ensuring efficient and scalable training. To regularize the model and prevent overfitting, we apply a dropout rate of 0.2. During the training process, we utilize the Adam optimizer with its default configuration, as outlined by [31], to optimize the model parameters and facilitate convergence towards an optimal solution.

### 4.3   Baseline Comparisons

We compare with SOTA baselines to thoroughly assess the performance of our model:

**Attention-based models**

1. ATAE-LSTM[2]: The attention mechanism directs the model's attention towards the most important part of the given sentence.
2. IAN[6]: An interactive attention mechanism and two LSTMs generate representations for both aspects and sentences.
3. RAM[3]: Employs a memory network with recurrent attention mechanisms to learn representations of sentences.
4. MGAN(Fan et al., 2018): The model learns how aspects and contexts interact by employing attention mechanisms that vary in granularity.
5. BERT[27]: The BERT model uses the input format "[CLS] sentence [SEP] aspect [SEP]" for processing text.
6. CBERT [22]: introduced conditional BERT contextual augmentation, a data augmentation technique specifically designed for labeled sentences in NLP.
7. AEN [12]: The paper introduces an Attentional Encoder Network (AEN), utilizing attention-based encoders to model the relationship between context and target without relying on recurrence.
8. IMAN [8]: The study proposed a novel approach utilizing Multi-head Attention (MHA) networks, combined with pre-trained embeddings from BERT, to enhance context-aspect interactions and improve ABSA.
9. MAN [1]: presented a Multi-Attention Network (MAN) addressing various challenges by employing intra-level attention with a transformer encoder and inter-level attention with global and local modules.
10. MAMN_W [23]: introduced a novel aspect-level sentiment analysis model leveraging bidirectional encoder representations from transformers (BERT)

      and multiple attention mechanisms, including intra- and inter-level attention, to enhance sentiment identification.

11. HN-PMAT[10]: The paper proposes a hybrid network model with shallow and deep layers and positional and interactive multi-head attention mechanisms to improve aspect-level sentiment classification.
12. IMHSACap[11]: IMHSACap incorporates Local Context Mask to emphasize local contexts and employs an interactive attention mechanism to capture long-range dependencies between global and local contexts, enhancing the model's understanding of aspect-related sentiment.

**Hybrid-based models**

13. RGAT+BERT [14]: RGAT builds upon the pre-trained BERT model to incorporate relation-aware graph attention mechanisms for enhanced performance.
14. BERT4GCN[15]: Combines the grammatical sequencing abilities of BERT's pre-trained language model with the syntactic understanding derived from dependency graphs.
15. TGCN[20]: It utilizes dependency types for distinguishing relations within the graph and employs an ensemble of attentive layers to capture contextual information across multiple GCN layers, enhancing the model's understanding of relationships and contextual nuances within the data.
16. AG-VSR+BERT[25]: knowledge through GCN.
17. KHGCN[17]: Introduced a novel approach for aspect-based sentiment analysis, utilizing a knowledge-guided heterogeneous graph convolutional network.
18. ASHGAT+BERT[18]: Developed an aspect-specific hypergraph attention network (ASHGAT) to analyze word-level relationships in aspect-based sentiment analysis, incorporating syntactic and semantic information and optimizing attention through aspect-oriented syntactic distance-based weighting.

Table 2. Experimental results comparison on three publicly available datasets

| Model | Restaurant14 | | Laptop14 | | Twitter | |
|---|---|---|---|---|---|---|
| | Acc. | F1 | Acc. | F1 | Acc. | F1 |
| ATAE-LSTM [23] | 77.20 | - | 68.70 | - | - | - |
| IAN [6] | 78.60 | - | 72.10 | - | - | - |
| RAM [3] | 80.23 | 70.80 | 74.49 | 71.35 | 69.36 | 67.30 |
| MGAN [9] | 81.25 | 71.94 | 75.39 | 72.47 | 72.54 | 70.81 |
| BERT [25] | 85.79 | 80.09 | 79.91 | 76.00 | 75.92 | 75.18 |
| CBERT [22] | 86.27 | 80.00 | 79.83 | 76.12 | 76.44 | 75.35 |
| AEN+BERT [12] | 83.12 | 73.76 | 79.93 | 76.31 | 74.71 | 73.13 |
| IMAN+BERT [8] | 83.95 | 75.63 | 80.53 | 76.91 | 75.72 | 74.50 |
| MAN [1] | 84.38 | 71.31 | 78.13 | 73.20 | 76.56 | 72.19 |
| MAMN_W [23] | 86.52 | **81.57** | 81.35 | 77.83 | 76.59 | 75.27 |
| HN_PMAT+BERT [10] | 85.13 | 76.21 | 79.71 | 75.80 | 75.45 | 73.30 |
| IMHSACap+BERT [11] | 85.00 | 77.90 | 81.03 | 77.62 | 76.30 | 75.19 |
| RGAT+BERT [14] | 86.60 | 81.35 | 78.21 | 74.07 | 76.15 | 74.88 |
| BERT4GCN [15] | 84.75 | 77.11 | 77.49 | 73.01 | 74.73 | 73.76 |
| TGCN+BERT [20] | 86.16 | 79.95 | 80.88 | 77.03 | 76.45 | 75.25 |
| AGVSR+BERT [24] | 86.34 | 80.88 | 79.92 | 75.85 | 76.45 | 75.04 |
| KHGCN+BERT [17] | - | - | 80.87 | 77.9 | - | - |
| ASHGAT+BERT [18] | 85.49 | 79.23 | 79.98 | 76.58 | - | - |
| A3SN (ours) | **86.86** | 80.85 | **81.96** | **78.37** | **77.10** | **75.94** |

### 4.4 Experimental Results

The overall performance of all the models is shown in Table 2, from which several observations can be noted. The A3SN model performs better than most baseline models across all three datasets, surpassing models incorporating semantic, syntactic, and knowledge information. The performance of BERT can be significantly improved when incorporated with A3SN, even without syntactic and external knowledge, making the model simpler. This proves that our A3SN is better at enhancing the model's ability to comprehend and leverage the inherent relationship between sentence and aspects, ultimately improving its performance. Third, proving the effectiveness of this massively pre-trained model for this purpose, the simple BERT can already outperform some of the current ABSA models by considerable margins. However, this robust model finds additional improvement by including our A3SN. The efficacy of our A3SN in capturing significant relationships between sentence and aspects for sentiment analysis has been proved by these results.

### 4.5   Ablation Study

Table 3. Results of an ablation study on three benchmark datasets (%).

| Model | Restaurant14 | Laptop14 | Twitter |
|---|---|---|---|
| | Acc. | Acc. | Acc. |
| A3SN | 86.86 | 81.96 | 77.10 |
| A3SN w/o original attention | 86.06 | 81.80 | 76.37 |
| A3SN w/o amplified attention | 86.33 | 81.85 | 76.66 |
| A3SN w/o gated fusion | 86.24 | 80.22 | 76.66 |

We performed ablation experiments on the datasets to examine the effects of various modules in our A3SN model on performance. The phrase "A3SN w/o original attention" describes how the representation "$head_{ori}^l$" obtained from the original attention mechanism in the textual semantic module has been removed. This entails replacing $head_{ori}^l$ and $Gate^O$ with $head_{amp}^l$ and $Gate^A$ in Eq. (11), respectively. Similarly, w/o amplified attention involves excluding the representation $head_{amp}^l$ calculated by the amplified attention mechanism in aspect-sentence awareness attention module, thereby using $head_{ori}^l$ and $Gate^O$ instead of $head_{amp}^l$ and $Gate^A$ in Eq. (11). Additionally, w/o gated fusion indicates the use of a fully connected network to integrate representations from the two modules without employing the fusion gate. The results are depicted in Table 3. Notably, without amplify aspect-sentence awareness attention, the performance of A3SN experiences a decrease of 0.53%, 0.11%, and 0.44% for the Restaurant, Laptop, and Twitter datasets, respectively. Furthermore, the amplified attention's representation in the aspect-sentence awareness attention module must be combined with the MHA's representation in the textual semantic module, as the performance of A3N decreases by 0.80%, 0.16%, and 0.73%, respectively, when solely relying on the amplified attention. Finally, A3SN performance drops by 0.62%, 1.70%, and 0.44% when a basic, fully connected network is substituted for the gated fusion module. Overall, A3SN performs better in capturing significant relationships between sentences and aspects for ABSA when all components are effectively combined. It can adaptively integrate two types of features from the textual semantic module and amplify the aspect-sentence awareness attention module into the transformer.

## 4.6 Case Study

To evaluate the efficacy of A3SN in capturing semantic information and the relationship between aspects and sentences for enhancing ABSA, we conducted a case study using a few sample sentences. Table 4 presents the predictions and corresponding truth labels for these sentences. In the second sample, "Our waiter was friendly, and it is a shame that he didn't have a supportive staff to work with," two aspects ("waiter" and "staff") with contrasting sentiment polarities are present, posing a challenge for existing attention models. A3SN adeptly determines the polarity of the aspect word "waiter" by integrating a textual semantic module for semantic comprehension of the aspect and the sentence and amplify the aspect-sentence awareness attention module to emphasize the relationship between the sentence and the input aspect "waiter." This doubled attention between the sentence and "waiter" enhances their relationship, reducing noise and emphasizing the significance of the "waiter" aspect over "staff." Finally, gated fusion dynamically assimilates valuable insights from both the textual semantic module representation and the aspect-sentence awareness attention module representation. The same process is applied if the input is "staff."

Table 4. Case studies of our A3SN model.

| Text | A3SN | Labels |
|---|---|---|
| Then the $[system]_{neg}$ would many times not $[power\ down]_{neg}$ without a forced power-off | (N, N) | (N, N) |
| Our $[waiter]_{pos}$ was friendly and it is a shame that he didn't have a supportive $[staff]_{neg}$ to work with. | (P, N) | (P, N) |
| Both a number of the $[appetizer]_{pos}$ and $[pasta\ specials]_{pos}$ were amazing. | (P, P) | (P, P) |
| $[Built-in\ apps]_{pos}$ are purely amazing. | (P) | (P) |
| It was our only opportunity to visit and wanted an authentic $[Italian\ meal]_{neu}$. | (O) | (O) |

### 4.7  Effect of A3SN Layer Number

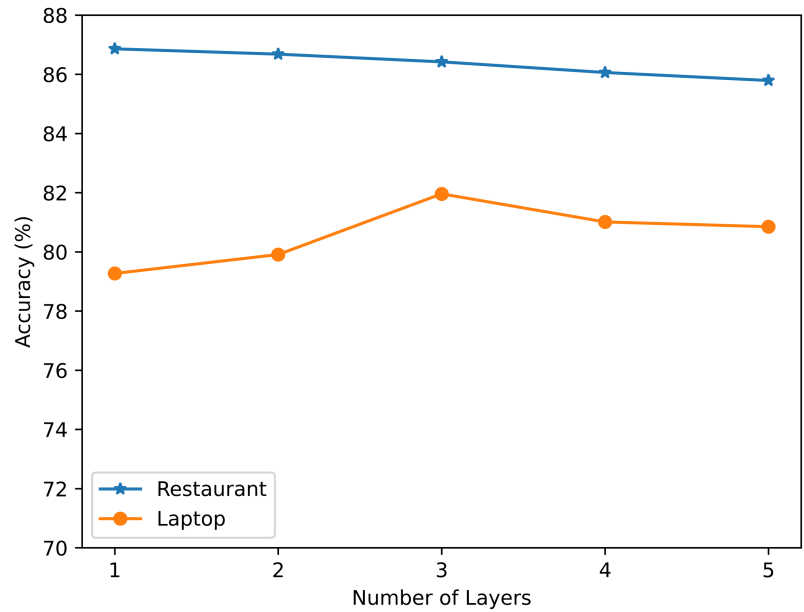

**Fig. 3.** Effect of different number of A3SN layers.

In our investigation, as depicted in Fig. 3, we observed a significant discrepancy in the optimal number of layers required for optimal model performance when applied to the Laptop and Restaurant datasets. Specifically, we found that the Laptop dataset yielded the best results with three layers, while the Restaurant dataset exhibited optimal performance with just one layer. This disparity in the number of layers needed for each dataset can be attributed to the inherent complexity and characteristics unique to each dataset. The Restaurant dataset, characterized by its relatively simpler structure and straightforward relationships between input sentences and corresponding aspects, lends itself well to effective modeling with fewer layers. In contrast, the Laptop dataset presents a more intricate landscape, necessitating deeper analysis and exploration of the relationships between sentences and aspects. The observed phenomenon underscores the importance of balancing model complexity and dataset characteristics. Employing too few layers on a complex dataset like Laptop can lead to underfitting. Conversely, an excessive number of layers can exacerbate overfitting.

### 4.8 Effect of Different A3SN Head Numbers

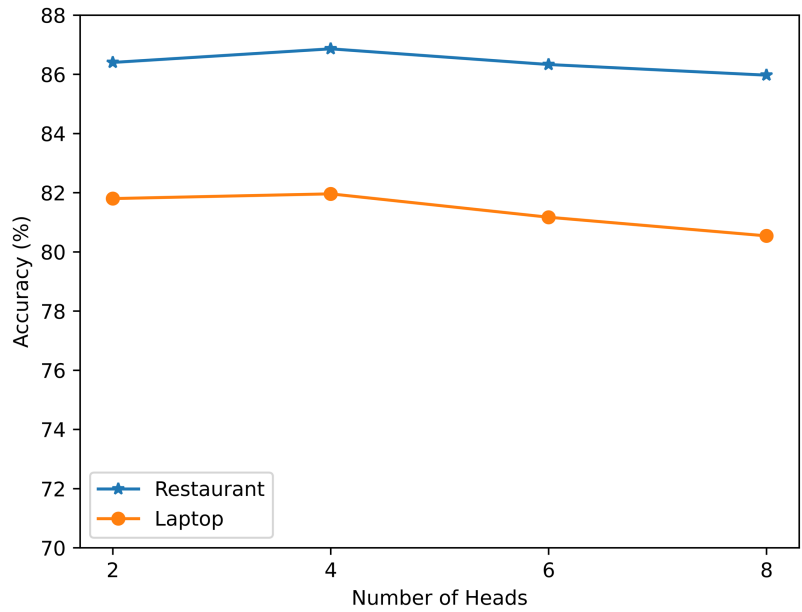

**Fig. 4.** Effect of different number of A3SN heads.

We illustrate with examples from the Restaurant and Laptop domains the effect of varying head numbers on the MHA mechanism of our models. As depicted in Fig. 4, optimal performance is observed when the head number is set to 4. This observation aligns with our intuition, suggesting that decreasing or increasing the head number results in a model that becomes prone to overfitting. This interpretation is grounded in the inherent characteristics of MHA, where an optimal balance is struck between complexity and generalization, leading to enhanced model performance.

## 5     Conclusion

In this work, we presented A3SN, a novel and efficient technique for enhancing ABSA. Our approach integrates MHA mechanisms to enrich the model with semantic information from the input text. We introduce an enhanced MHA mechanism that specifically amplifies aspect-sentence awareness attention to captures subtle relationships and dependencies between aspects and their contextual sentences. The gated fusion mechanism further integrates these feature representations. Experimental results on three benchmark datasets confirm A3SN's superior performance over state-of-the-art baseline models, achieving remarkable effectiveness while maintaining simplicity and low computational complexity. These results validate the robustness and efficiency of A3SN, making it a valuable advancement in sentiment analysis.

**Funding.** The Chinese government scholarship supported this work.

**Data availability**. Data will be made available at a reasonable request.

**Disclosure of Interests.** The authors have no competing interests to declare relevant to this article's content.

24       A. Lawan et al.